\documentclass[10pt,conference,a4paper]{IEEEtran}
\ifCLASSINFOpdf
  \usepackage[pdftex]{graphicx}
\else
  \usepackage[dvips]{graphicx}
\fi
\usepackage{amsmath}
\usepackage{algorithmic}

\usepackage{array}

\ifCLASSOPTIONcompsoc
  \usepackage[caption=false,font=normalsize,labelfont=sf,textfont=sf]{subfig}
\else
  \usepackage[caption=false,font=footnotesize]{subfig}
\fi
\usepackage{url}

\usepackage{textcomp}
\usepackage{bm}
\usepackage{multirow}

\def\eg{\emph{e.g}}

\def\etal{\emph{et al}}

\begin{document}
%
\title{Recycle deep features for better object detection}

\author{\IEEEauthorblockN{Wei Li \qquad \qquad Matthias Breier \qquad \qquad Dorit Merhof}
\IEEEauthorblockA{Institute of Imaging and Computer Vision, RWTH Aachen University\\
52074 Aachen, Germany\\
Email: wei.li@lfb.rwth-aachen.de}}


%


\maketitle

\begin{abstract}
Aiming at improving the performance of existing detection algorithms developed for different applications, we propose a region regression-based multi-stage class-agnostic detection pipeline, whereby the existing algorithms are employed for providing the initial detection proposals. Better detection is obtained by exploiting the power of deep learning in the region regress scheme while avoiding the requirement on a huge amount of reference data for training deep neural networks. Additionally, a novel network architecture with recycled deep features is proposed, which provides superior regression results compared to the commonly used architectures. As demonstrated on a data set with \texttildelow{1200} samples of different classes, it is feasible to successfully train a deep neural network in our proposed architecture and use it to obtain the desired detection performance. Since only slight modifications are required to common network architectures and since the deep neural network is trained using the standard hyperparameters, the proposed detection is well accessible and can be easily adopted to a broad variety of detection tasks.
\end{abstract}


%
\IEEEpeerreviewmaketitle

\section{Introduction}
\label{sec:intro}
Object detection is one of the prime tasks in computer vision. In general, there are two paradigms for seeking objects in images. The first and probably the most widely used paradigm is to manually define some distinctive features for localizing the objects of interest, where some prior knowledge-based heuristics are employed to separate them from background regions. Many conventional text detection algorithms~\cite{Gatos2005,Epshtein2010,Yi2011} fall into this class. Obviously, such methods are only applicable for well defined problems with low variations of objects and background. If the sought objects and the presented scenes exhibit a great diversity, the alternative detection paradigm is often applied, where machine learning-based analysis is involved. In comparison to hand-crafted features and heuristic rules, the most distinctive features are extracted in images and appropriate separation boundaries are constructed in the corresponding feature space by automatically exploiting the information embedded in learning samples. Some recent research~\cite{Erhan2014,Redmon2015} has shown its success using whole images as the input to carefully designed deep neural networks~({DNNs})~\cite{LeCun2015} and provide an end-to-end detection solution merely based on learning data. However, the successful training of such detectors requires a huge number of images with full annotations to all objects, which leads to a challenging data problem if multi-class objects are simultaneously presented in images. Indeed, the amount of required training data can be reduced if the detection process is decomposed into multiple stages. As demonstrated in~\cite{Felzenszwalb2010,Girshick2016}, after generating detection proposals using either exhaustive search or dedicated segmentation, detectors are trained to reject false alarms while retaining true objects. For each image, multiple positive and negative samples can be obtained and used for the training purpose.

The remarkable work from Krizhevsky~\etal.~\cite{Krizhevsky2012} demonstrates the practical application of deep learning ({DL}) on large-scale data problems and its superior performance in comparison to conventional methods. Currently, break-throughs have been achieved in many fields~\cite{Ciregan2012,Sak2014,Szegedy2015} using {DL}. Among the variants of deep architectures, convolutional neural networks~({CNNs}) have been widely used due to their computational efficiency in training.

In our work, we adopted the deep learning-based multi-stage detection paradigm and demonstrate its advantage in the case of limited training data. Our idea is to use the well established detection/segmentation algorithms existing for diverse applications and improve the detection performance by exploiting the power of {DL}. In such a manner, the previously obtained prior knowledge about the application is used to alleviate the training of {DNNs} since the diversity of the test objects into {DNNs} is substantially reduced. As a result, some generic architectures of DL for dealing with object detection on the basis of reasonable region proposals are realizable. In comparison to conventional training strategies, we only consider positive samples of objects in a region regression scheme while avoiding the definition of negative samples, which is sometimes non-trivial. Even without a dedicated classification, good detection results can be obtained using the newly introduced post-regression-overlapping-pre-regression ($PoP$) score. The second major contribution of our work is the novel {CNN} architecture for improving the regression performance. By recycling deep features from lower scales, more comprehensive information is employed to predict the position and the size of the sought object with a higher precision. As a desirable side effect of this architecture, more stable convergence in training is obtained and {CNNs} can be trained using the standard hyperparameters, which makes the proposed detection more accessible for users.

The remaining of this paper is organized as follows: after describing the proposed object detection in the region regression scheme in Section~\ref{sec:detection_regression}, we introduce in Section~\ref{sec:recycled_features} our novel {CNN} architecture with recycled deep features. To investigate the performance of the proposed architecture, we apply it in comparison to commonly used architectures for detecting objects on printed circuit boards ({PCBs}) and give a comprehensive evaluation in Section~\ref{sec:performance}. For the reader's convenience, a general conclusion is drawn in Section~\ref{sec:conclusion}.

\section{Object Detection Using Region Regression}
\label{sec:detection_regression}
There are two dominant error modes in machine learning-involved multi-stage object detection: the miss-classification of region proposals and the inaccurate localization of objects. Thanks to the highly discriminative classification using {DL}, errors of the first group can be significantly suppressed. To deal with the poor localization, the common solution is generating adequate region proposals of sought objects and using some pre-trained predictors to re-localize the objects to more accurate positions in region proposals~\cite{Jaderberg2015,Girshick2016}.

In general, the objects of interest are well covered by the training samples. However, the variation of background in real detection is not necessarily well represented in the provided data sets. This raises the question how negative samples should be defined for achieving a reliable classification between true and false candidates. To solve this problem, our suggestion is to train {CNNs} using only positive samples and expect a reasonable response only in case of true objects. This is even advantageous regarding the reduced diversity of samples and the amount of required training data also decreases. To bridge the gap between the intended discrimination between two classes and the employed one-class training data, the region regression scheme is applied in combination with the $PoP$ score for canceling false alarms.

Ideally, all region proposals are represented using bounding boxes tightly enclosing the sought objects and we do not distinguish between proposals from different classes of objects. Let $({x_{1}},\, {y_{1}})$ and $({x_{2}},\, {y_{2}})$ denote the upper-left and lower-right corners of the input bounding box in the corresponding image, respectively. In a similar way, $({x_{1p}},\, {y_{1p}},\, {x_{2p}},\, {y_{2p}})$ denote the coordinates of the predicted bounding box after region regression. To remove the dependency of the predicted bounding box on the size of the detected object, the regression result $({\tilde{x}_{1p}},\, {\tilde{y}_{1p}},\, {\tilde{x}_{2p}},\, {\tilde{y}_{2p}})$ is always normalized as
\begin{equation}
	{\tilde{x}_{1p/2p}} = \frac{{x_{1p/2p}} - {c_{x}}}{w},\quad {\tilde{y}_{1p/2p}} = \frac{{y_{1p/2p}} - {c_{y}}}{h},
\label{eq:bb_normalized}
\end{equation}
where
\begin{equation}
	{c_{x}} = \frac{{x_{1}} + {x_{2}}}{2} ,\, {c_{y}} = \frac{{y_{1}} + {y_{2}}}{2} ,\, w = {x_{2}} - {x_{1}} ,\, h = {y_{2}} - {y_{1}}. \nonumber
\end{equation}
Inverting Eq.~\ref{eq:bb_normalized}, the absolute coordinates of the predicted bounding box are obtained given the regression result $({\tilde{x}_{1p}},\, {\tilde{y}_{1p}},\, {\tilde{x}_{2p}},\, {\tilde{y}_{2p}})$:
\begin{equation}
	{x_{1p/2p}} = {\tilde{x}_{1p/2p}} \cdot w + {c_{x}},\quad {y_{1p/2p}} = {\tilde{y}_{1p/2p}} \cdot h + {c_{y}}.\\
\label{eq:bb_absolute}
\end{equation}

For any positive sample with its ground-truth bounding box $({x_{1t}},\, {y_{1t}},\, {x_{2t}},\, {y_{2t}})$, the expected region regression output $({\tilde{x}_{1t}},\, {\tilde{y}_{1t}},\, {\tilde{x}_{2t}},\, {\tilde{y}_{2t}})$ is calculated according to Eq.~\ref{eq:bb_normalized} with $({x_{1p}},\, {y_{1p}},\, {x_{2p}},\, {y_{2p}}) = ({x_{1t}},\, {y_{1t}},\, {x_{2t}},\, {y_{2t}})$. If the regression result $({\tilde{x}_{1p}},\, {\tilde{y}_{1p}},\, {\tilde{x}_{2p}},\, {\tilde{y}_{2p}})$ is obtained for any input region proposal, the absolute coordinates $({x_{1p}},\, {y_{1p}},\, {x_{2p}},\, {y_{2p}})$ of the predicted bounding box are recovered according to Eq.~\ref{eq:bb_absolute}. To determine the parameters of the CNN for achieving the desired regression performance, it is important to define an appropriate loss function for guiding the training procedure. We define the loss $F$ as the mean squared distance between the expected and obtained regression results:
\begin{equation}
	F = \frac{1}{4 \cdot L} \sum^{L}_{l=1} ({\bm{r}_{l,t}} - {\bm{r}_{l,p}})^{T} ({\bm{r}_{l,t}} - {\bm{r}_{l,p}}),
\label{eq:loss}
\end{equation}
where $l$ indicates the $l$-th of all $L$ training samples. ${\bm{r}_{l,t}}$ and ${\bm{r}_{l,p}}$ are the normalized ground-truth and predicted bounding boxes of the $l$-th sample, respectively. Finally, the optimal network parameters are determined by minimizing $F$.

For canceling false alarms, indicative values should be provided for all region proposals and be used to quantify the confidence of being a sought object. We consider the consistency between the bounding boxes before and after the region regression as a good indicator and thus propose the use of the $PoP$ score
\begin{equation}
	PoP = \frac{({x_{1}},\, {y_{1}},\, {x_{2}},\, {y_{2}}) \cap ({x_{1p}},\, {y_{1p}},\, {x_{2p}},\, {y_{2p}})}{({x_{1}},\, {y_{1}},\, {x_{2}},\, {y_{2}}) \cup ({x_{1p}},\, {y_{1p}},\, {x_{2p}},\, {y_{2p}})}
\label{eq:pop_score}
\end{equation}
for assessing all detection proposals, where $\cap$ and $\cup$ denote the intersection and the union between two regions, respectively. The idea of the $PoP$ score is quite straightforward: if any object is well captured by the current bounding box, the regression output must be similar to the input. If an insufficient proposal or background is presented, the deployed CNN attempts to significantly change the bounding box to find more reasonable region. The difference could become more traceable if the region regression iterates multiple times.

Since our detection pipeline is designed for generic purposes, it can be easily adopted for improving the detection performance in diverse applications. Using existing detection algorithms for generating region proposals, the CNN for conducting region regression can be trained using a data set of moderate size and subsequently deployed to improve the detection rate and the localization precision of the proposals. Moreover, the cancellation of false detection proposals can be simultaneously realized using the regression results and regarding the resulting $PoP$ score. It is also possible to feed the obtained region proposals to dedicated classifier for achieving further improved detection performance.

\begin{figure*}[!t]
\centering
\subfloat[architecture for classification]{\includegraphics[width=0.4\linewidth]{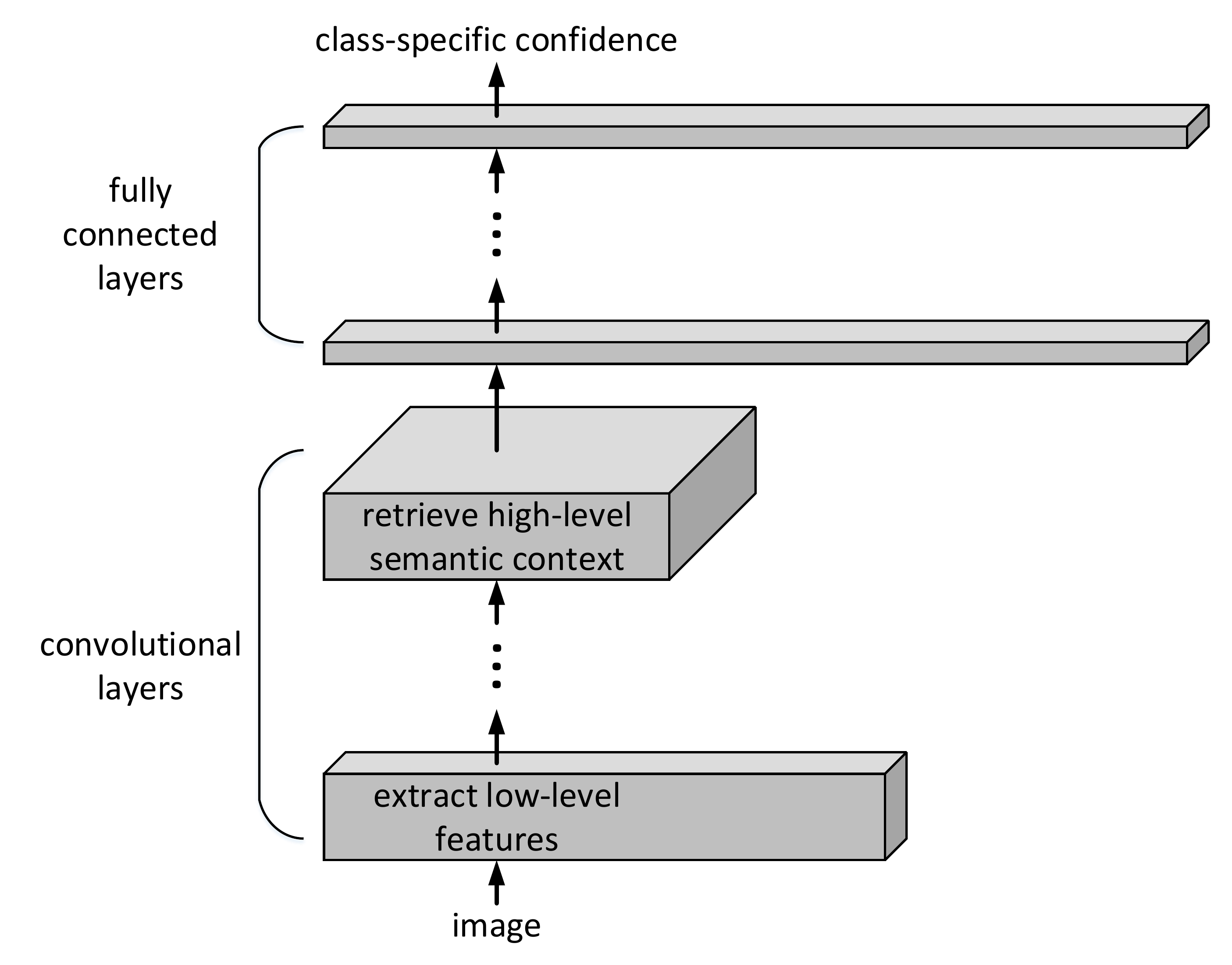}%
\label{fig:architecture_classification}}
\hfil
\subfloat[architecture for region regression]{\includegraphics[width=0.4\linewidth]{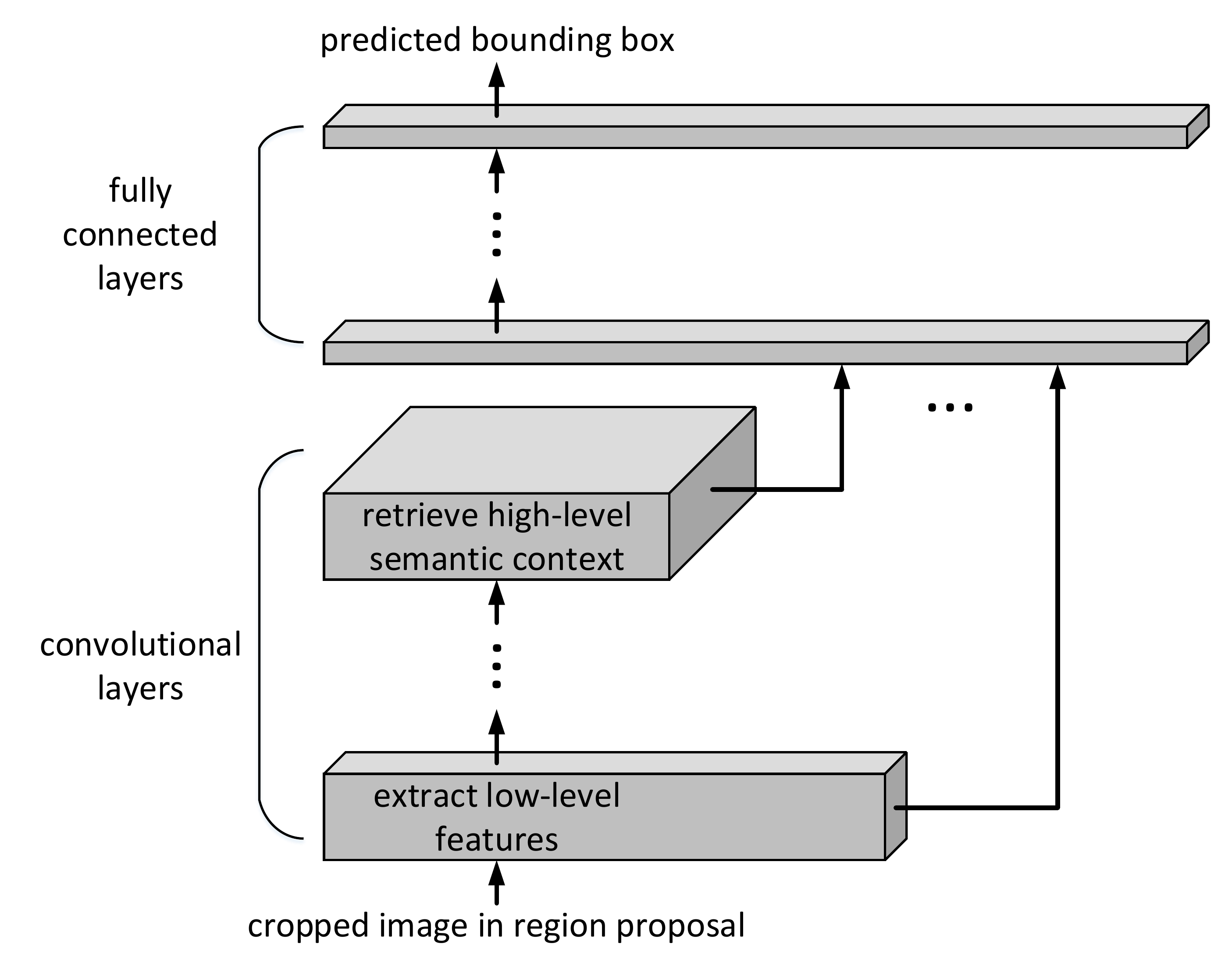}%
\label{fig:architecture_regression}}
\caption{CNN architectures for different tasks.}
\label{fig:architectures}
\end{figure*}

\section{Region Regression with Recycled Deep Features}
\label{sec:recycled_features}
So far, we have considered the high-level design of the proposed detection pipeline, but the appropriate architecture of the deployed CNN for region regression is still an open question. In literature, region regression is implemented either using the discriminative deep features in the sense of classification~\cite{Girshick2016} or adopting CNN architectures primarily for classification~\cite{Erhan2014,Jaderberg2015}. However, we have concerns about the applicability of such implementations in the case of region regression since detection is in general different from classification.

As illustrated in Fig.~\ref{fig:architecture_classification}, the common CNN architectures for classification purposes follow the idea of extracting low-level features at bottom layers and retrieving high-level semantic context at top layers. The output of the top convolutional layer is fed to fully connected layers (fcs), which construct the complex separation boundaries between classes. Such architectures result in classification relying on context information with high invariance instead of relying on less reliable low-level features~\cite{Zeiler2014}, \eg. gradient, texture, single discriminative parts. In contrast, in the case of region regression, the localization precision can benefit from the spatial information of low-level features since the position and the size of the sough object is retrievable according to the spatial distribution of these features. As complementary information, the high-level context determines the appropriate regression function for the presented object and avoids extreme predictions.

\begin{table}[!b]
\renewcommand{\arraystretch}{1.0}
\caption{Statistics of PCB components.}
\vspace{-0.5cm}
\label{tab:data}
\begin{center}
\begin{tabular}{|l|c|c||l|c|c|}
\hline
\multicolumn{1}{|c|}{\multirow{2}{*}{class}} & \multicolumn{1}{c|}{\multirow{2}{*}{samples}} & mean size & \multicolumn{1}{c|}{\multirow{2}{*}{class}} & \multicolumn{1}{c|}{\multirow{2}{*}{samples}} & mean size\\
\multicolumn{1}{|c|}{} & \multicolumn{1}{c|}{} & (in pixels) & \multicolumn{1}{c|}{} & \multicolumn{1}{c|}{} & (in pixels)\\
\hline
battery & 9 & 83292 & capacitor & 363 & 11075\\
\hline
CPU & 7 & 777736 & cooling & 22 & 399066\\
\hline
diode & 11 & 8034 & connector & 38 & 98216\\
\hline
IC & 203 & 23258 & inductor & 45 & 23553\\
\hline
LED & 6 & 5105 & oscillator & 33 & 10860\\
\hline
slot & 318 & 69650 & transistor & 139 & 11325\\
\hline
other & 30 & 60360 & \multicolumn{3}{c}{}\\
\cline{1-3}
\end{tabular}
\end{center}
\end{table}

Apparently, the classification {CNN} architectures are unable to maintain all necessary information for achieving the desired localization performance. To overcome this drawback of standard architectures and to avoid any potential problems caused by strongly alternating the well established CNNs, we propose a minimal but effective modification to the existing CNNs, whereby the novel architecture ``CNN with recycled deep features" (CNN-WRDF) for region regression is defined. For a better understanding, the generic construction of CNN-WRDF is illustrated in Fig.~\ref{fig:architecture_regression}. In addition to the output from the top convolutional layer, all intermediate features extracted in lower convolutional layers are also forwarded to fcs for providing the necessary spatial information. Interestingly, recycling deep features from lower layers brings a very desirable side effect: the path for passing information between different layers becomes shorter! We can consider the training of CNNs as a control process for achieving the desired system output, which is in fact the ground-truth bounding boxes for the input region proposals. The controller is in our case the optimizer minimizing the loss $F$ and determining the optimal network parameters. Using appropriate optimization methods, typically stochastic gradient descent~\cite{Bottou2012} or Nesterov{\textquotesingle}s accelerated gradient~\cite{Sutskever2013}, network parameters are updated in an iterative manner for decreasing the error between the desired and the currently obtained output. It is well known that if the dead time of the target system is significant, it could be difficult for the system to reach the desired stable state. Similarly, due to the deep architecture in CNNs, there is a long path for passing information between bottom and top layers in the training stage, which could lead to difficulties in reaching stable convergence during optimization. With the recycled deep features, additional forward paths from lower layers to top layers are introduced for strengthening their correlations.

\begin{table}[!b]
\renewcommand{\arraystretch}{1.0}
\caption{Comparison between proposal generation algorithms.}
\vspace{-0.5cm}
\label{tab:proposal_comparison}
\begin{center}
\begin{tabular}{|c|c|c|c|c|}
\hline
& Selective Search & BING & Edge Boxes & MST\\
\hline
proposals & 1182745 & 95279 & 297590 & 90141\\
\hline
matches & 1221 & 304 & 1075 & 1084\\
\hline
recall & 0.998 & 0.248 & 0.878 & 0.886\\
\hline
\end{tabular}
\end{center}
\end{table}

\section{Performance investigation}
\label{sec:performance}

\begin{table*}[!t]
\renewcommand{\arraystretch}{1.0}
\caption{Layers of the basis CNN.}
\vspace{-0.5cm}
\label{tab:standard_cnn}
\begin{center}
\begin{tabular}{|c|c|c|c||c|c|c|c|}
\hline
layer & type & kernel size/stride & output & layer & type & kernel size/stride & output\\
\hline
1 - 2 & conv + relu & (3, 3)/1 - (1, 1)/1  & (128, 128, 32) & 13 - 14 & conv + relu & (3, 3)/1 - (1, 1)/1 & (16, 16, 192)\\
\hline
3 - 4 & conv + relu & (3, 3)/2 - (1, 1)/1 & (64, 64, 32) & 15 & max pool & (3, 3)/2 & (8, 8, 192)\\
\hline
5 & max pool & (3, 3)/2 & (32, 32, 32) & 16 - 17 & conv + relu & (3, 3)/1 - (1, 1)/1 & (8, 8, 256)\\
\hline
6 - 7 & conv + relu & (3, 3)/1 - (1, 1)/1 & (32, 32, 64) & 18 - 19 & conv+relu & (3, 3)/1 - (1, 1)/1 & (8, 8, 384)\\
\hline
8 - 9 & conv + relu & (3, 3)/1 - (1, 1)/1 & (32, 32, 64) & 20 & fc + relu + dropout & - & (1024)\\
\hline
10 & max pool & (3, 3)/2 & (16, 16, 64) & 21 & fc + dropout & - & (1024)\\
\hline
11 - 12 & conv + relu & (3, 3)/1 - (1, 1)/1 & (16, 16, 128) & 22 & fc & - & (4)\\
\hline
\end{tabular}
\end{center}
\end{table*}

\subsection{Data Specification}
\label{sec:perf_data}
To investigate the performance of the proposed detection pipeline and the significance of the novel CNN-WRDF architecture, we conducted a series of evaluations on the PCB data set~\cite{Li2016}, which consists of 31 images from different PCBs and presents in total 1224 fully annotated components. Some statistics of this data set are listed in Table~\ref{tab:data}. As stated in Section~\ref{sec:detection_regression}, adequate region proposals of sought objects should be generated in the first place using some conventional detection/segmentation algorithms. Regarding this, we employed a simple multi-scale thresholding algorithm (MST) to obtain the initial bounding boxes for the CNN-based region regression. A comparison between MST and state-of-the-art proposal generation algorithms~\cite{Uijlings2013,Cheng2014,Zitnick2014} on the PCB data set is presented in Table.~\ref{tab:proposal_comparison}. For the convenience of using CNNs, all region proposals were warped to the size $128 \times 128$.

In consideration of separated training and test stages, the 31 images were divided into two parts: a training set of 25 images and a test set of the other 6 images. Obviously, it is challenging to train a regression CNN using data of such a limited size. Thus, we first managed to enlarge the number of the training samples. On the one hand, instead of assigning maximally one proposal to each ground-truth object, all proposals with an intersection-over-union (IoU) score over 0.5 with any components were considered as valid positive samples for training. On the other hand, all ground-truth objects and the positive proposals generated using a standard segmentation algorithm~\cite{Felzenszwalb2004} were also included in the training set. All of these samples were augmented by some predefined sizes for obtaining further new data. Moreover, similar to the approach applied in \cite{Krizhevsky2012}, additional training data were generated using spatial transformations: rotating each sample over different angles and flipping it with respect to the x and y-axes. In the end, a total of 358704 samples was provided for training CNNs. It should be emphasized that only the training data were employed for analyzing the applicability of CNNs if necessary, and the test data were merely used in the test stage for assessing detection performance.

\subsection{Network Architectures and Training}
\label{sec:perf_nets_training}
Towards a comprehensive analysis of the proposed architecture with recycled deep features, we evaluated it in combination with diverse CNNs. The first variant is a basis CNN for classification tasks and all necessary modifications were made for applying it in the context of region regression. The detailed layer organization can be found in Table.~\ref{tab:standard_cnn}. This CNN serves as the basis architecture in our evaluation and all other architectures are realized by extending it using different methods. Besides the proposed CNN-WRDF architecture, we are also interested in the approach ``Network in Network"~\cite{Lin2013} (NIN) for better model discriminability, in dropout~\cite{Hinton2012} and dimension reduction\cite{Szegedy2015} for more reliable features, and also in additional information, \eg. the absolute size of bounding boxes, for potentially improving the regression performance.

For training CNNs, we employed the DL framework ``Convolutional Architecture for Fast Feature Embedding"~\cite{Jia2014} (Caffe) and used the NESTEROV~\cite{Sutskever2013} solver for optimizing network parameters. Since good accessibility of CNNs for users even without much experience with DL is desired, we set all training parameters to the standard values: the inverse learning rate policy with the base learning rate 0.01 and the learning decay 0.90 for each epoch, the momentum value 0.9 and the weight decay 0.0005~\cite{Krizhevsky2012}. Especially, considering the capacity of the mainstream GPU devices, the batch size was set to 64, which results in a requirement on memory below 4 GB. Correspondingly, the total number of iterations was set to 200000 for about 36 epochs.

\subsection{Detection Strategy}
\label{sec:perf_strategy}
The most important strategy applied in our detection pipeline is augmenting the corresponding region of proposals before feeding them into CNNs. This is inspired by the strategy involved in natural object detection and recognition: objects are better localized and recognized if the context information in the neighborhood region is also considered. Thus, to also include the neighborhood context information for those region proposals only partly covering the corresponding objects, we augmented all region proposals by a certain size proportional to their original dimensions. Recalling that our CNNs are only trained using positive samples, improved discriminability between true and false detection proposals can be expected if this strategy is combined with the $PoP$ score after iterative regression, because for alternating size of false alarms it is unlikely to obtain a high consistency between the original and the predicted bounding boxes.

\begin{figure*}[!t]
\centering
\subfloat[]{\includegraphics[width=0.42\linewidth]{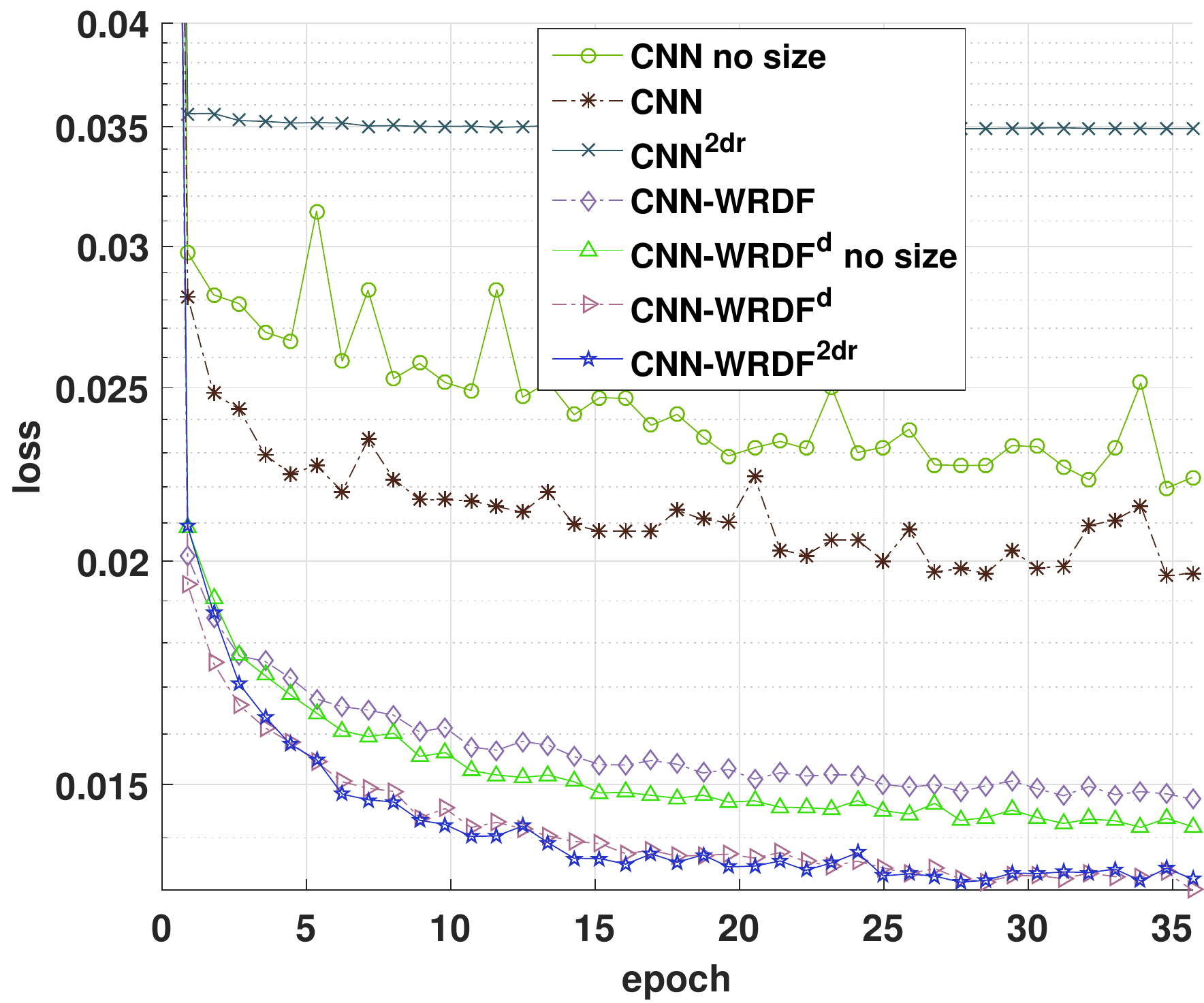}%
\label{fig:comparison_comprehensive}}
\hfil
\subfloat[]{\includegraphics[width=0.42\linewidth]{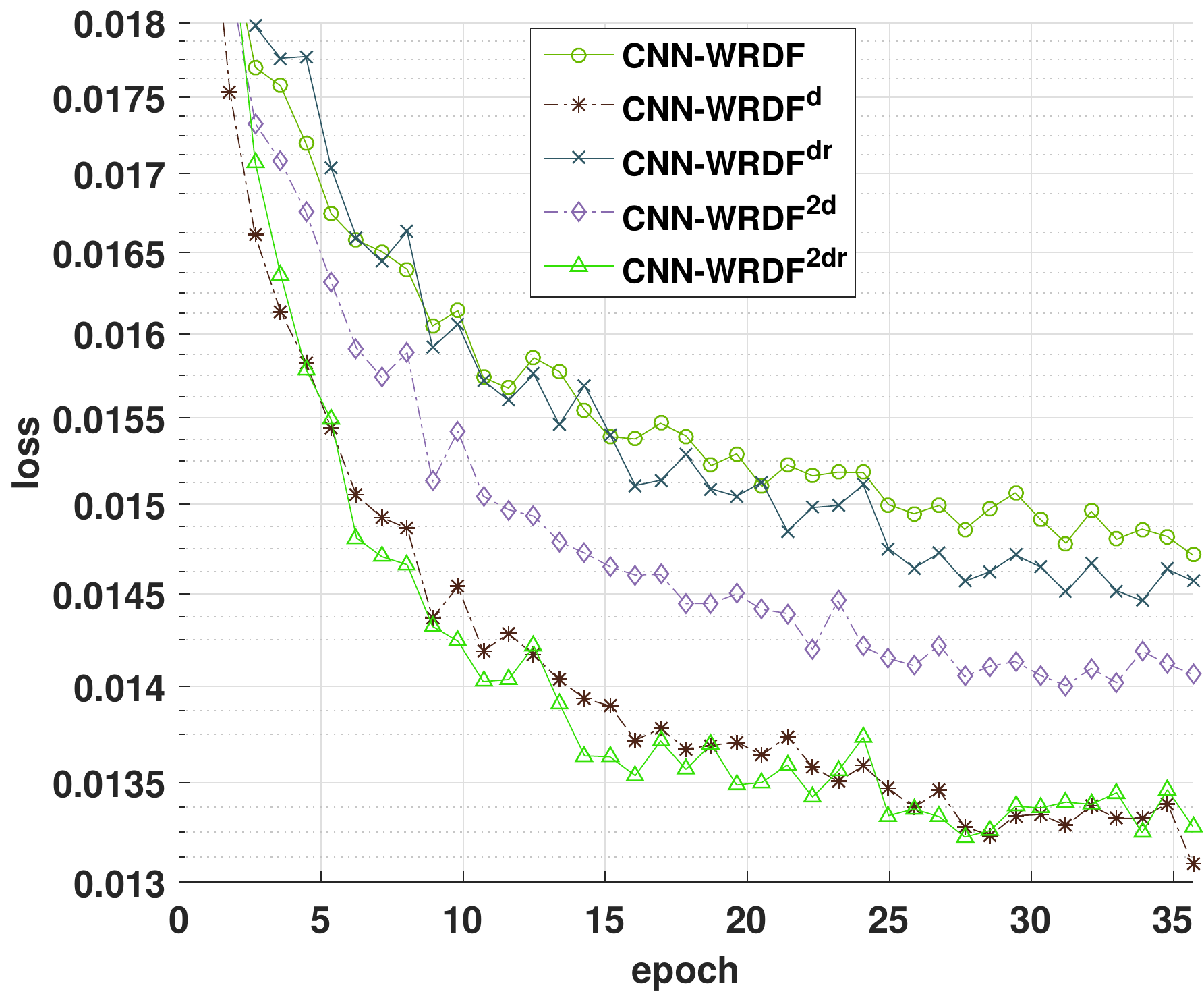}%
\label{fig:comparison_recycling}}
\caption{Comparison between different CNN architectures. The basis CNN is denoted with ``CNN". Any CNNs extended with the recycled deep features are denoted with the suffix ``-WRDF". ``no size" denotes the absence of the absolute size of bounding boxes in CNNs. The superscript ``2" denotes the use of the original NIN with two cascaded $1 \times 1$ convolutional cross channel layers instead of the only one cross channel layer in the basis CNN. The other two superscripts ``d" and ``r" denote the additional dropout and dimension reduction of the forwarded features, respectively.}
\label{fig:comparison}
\end{figure*}

\subsection{Results}
\label{sec:perf_results}
The first evaluation was on the feasibility of training CNNs without the spatial transformation-based data augmentation in Section~\ref{sec:perf_data}. None of the considered CNNs was successfully trained using the original data, while stable convergence of the loss $F$ has been observed during training the same CNNs using the augmented data.

Then, we trained all CNNs under the same conditions and using the same hyperparameters. In the test stage, their regression performance was assessed on the test data and used to investigate the significance of different extensions to the basis architecture. Representative results are visualized in Fig.~\ref{fig:comparison} for highlighting some generic statements. For convenience, the basis CNN is denoted with ``CNN". Any CNNs extended with the recycled deep features are denoted with the suffix \mbox{``-WRDF"}. ``no size" denotes the absence of the absolute size of bounding boxes in CNNs. The superscript ``2" denotes the use of the original NIN with two cascaded $1 \times 1$ convolutional cross channel layers instead of the only one cross channel layer in the basis CNN. The other two superscripts ``d" and ``r" denote the additional dropout and dimension reduction of the forwarded features, respectively. In Fig.~\ref{fig:comparison_comprehensive}, superior results are obtained if our proposed CNN-WRDF architecture is applied. Slight further improvement of the performance can be achieved by considering the absolute size of bounding boxes. In Fig.~\ref{fig:comparison_recycling}, better results are obtained if the dropout of the forwarded features is included in CNNs. In contrast, no definite benefits can be observed if the standard NIN and the dimension reduction of features are included in CNNs.

\begin{figure*}[!t]
\centering
\subfloat[training data]{\includegraphics[width=0.42\linewidth]{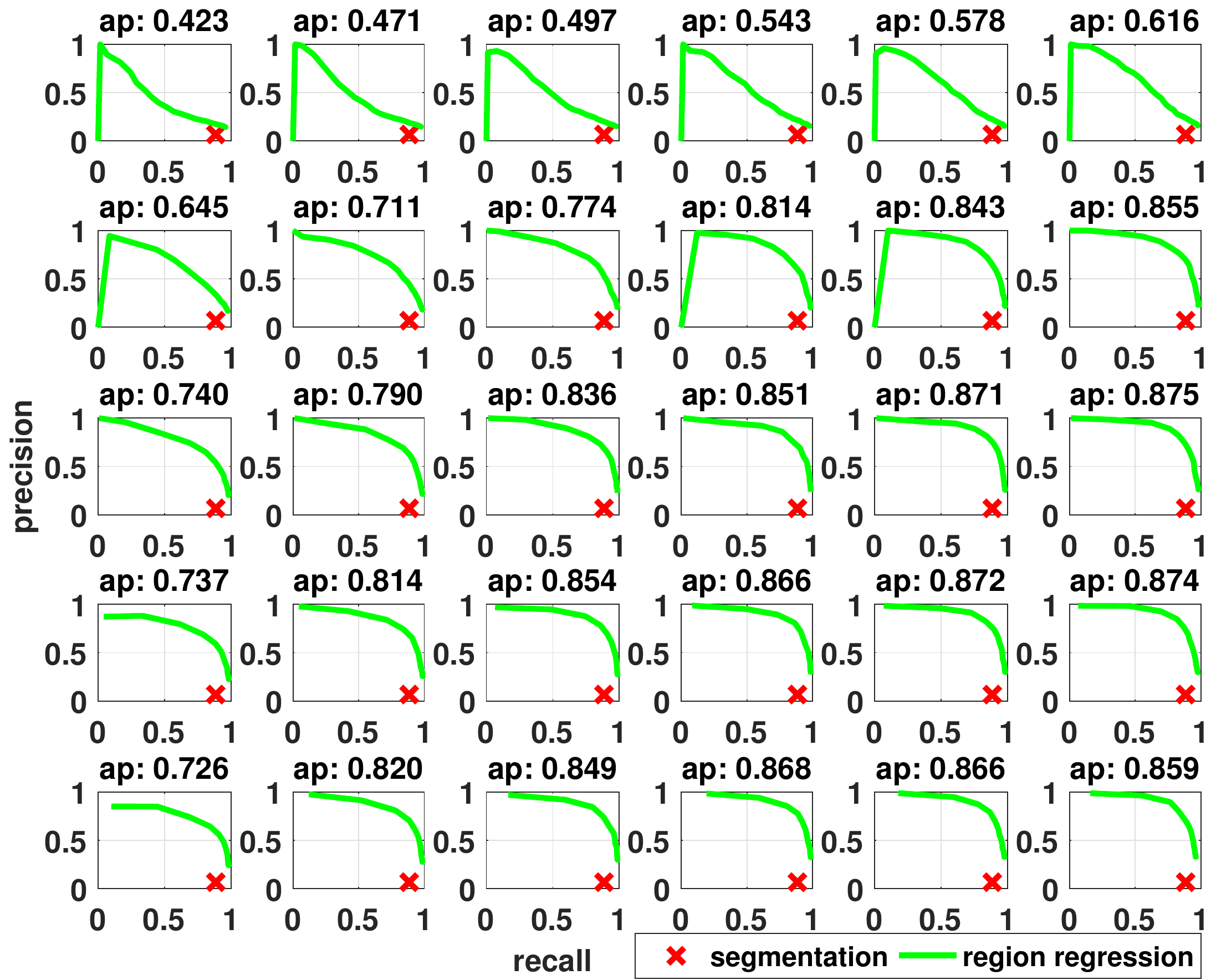}%
\label{fig:rc_curves_training}}
\hfil
\subfloat[test data]{\includegraphics[width=0.42\linewidth]{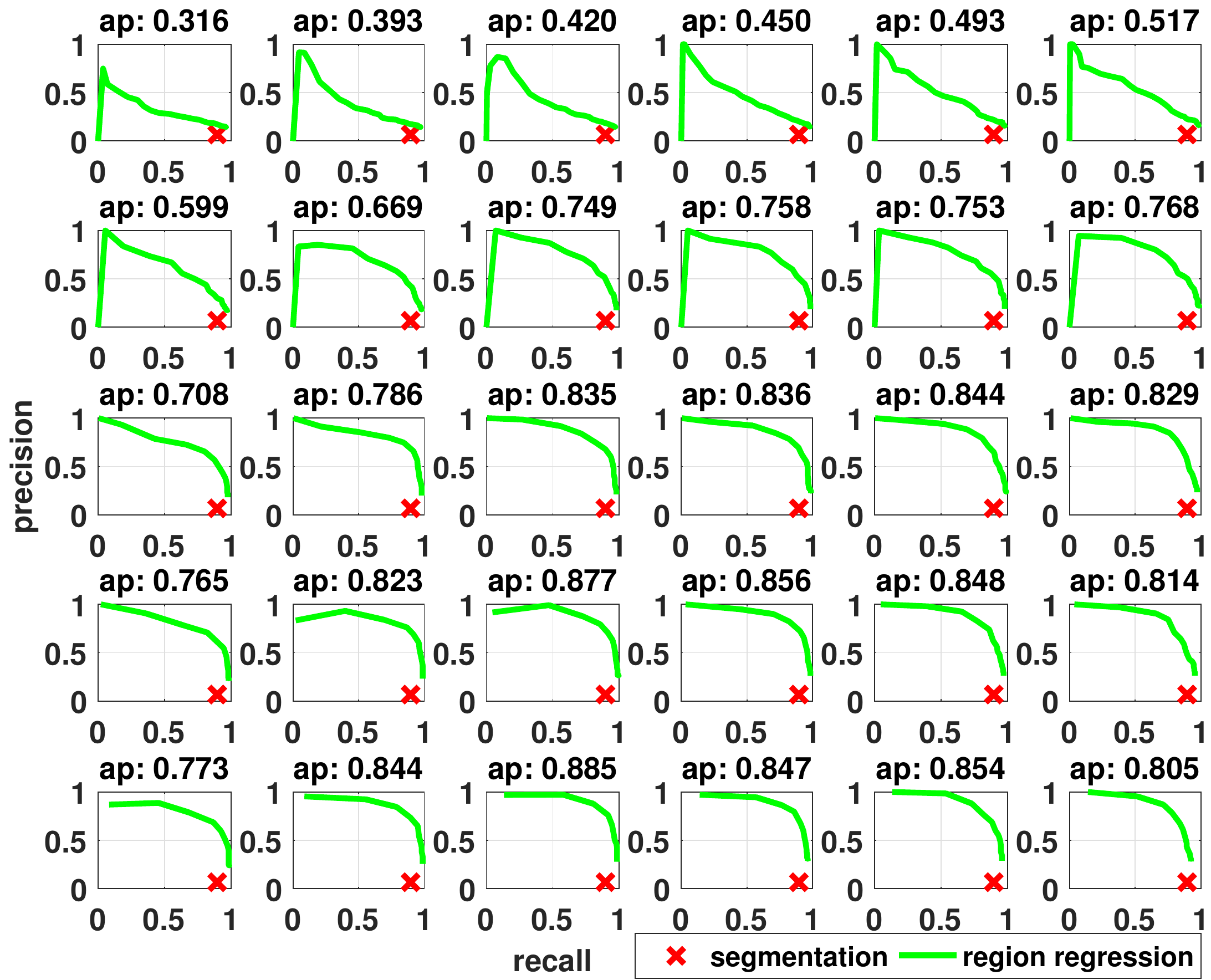}%
\label{fig:rc_curves_test}}
\caption{Detection for different parameter settings of augmentation size and iteration number. The columns from left to right are with the augmentation sizes: 0, 0.1, 0.2, 0.3, 0.4, 0.5, respectively. The rows from top to bottom are with the iteration numbers: 1, 2 3, 4, 5, respectively. All precision-recall plots are obtained by decreasing the threshold of the $PoP$ score from 1 to 0. The training and test data are considered separately.}
\label{fig:rc_curves}
\end{figure*}

Finally, the CNN with recycled deep features and the dropout of forwarded features (CNN-WRDF$^{\text{d}}$) was deployed in detection and the $PoP$ score was calculated for distinguishing between true and false proposals. As stated in Section~\ref{sec:perf_strategy}, it could be better to use the augmented region proposals as input to the CNN-based regression and predict the bounding boxes of sought objects in an iterative manner. To investigate the impact of the augmentation size and the number of iterations, detection results were obtained for the augmentation sizes from 0 to 0.5 with the step size of 0.1 and for the iteration numbers from 1 to 5 with the step size of 1. An overview of the corresponding detection performance is presented in Fig.~\ref{fig:rc_curves}. For a better visualization, precision-recall curves are plotted by decreasing the threshold of the $PoP$ score from 1 to 0. The average precision ($ap$) score is also provided for the convenience of comparison. Generally, the augmentation strategy combined with iterative regression significantly improves the detection performance. However, the selected augmentation size should be appropriate. Otherwise, high iteration numbers could lead to a decrease in the quality of detection. Furthermore, regarding the similar performance obtained on the training and test data separately, it is possible to determine the optimal combination between the augmentation size and the iteration number in training, which leads to satisfactory detection in test.

\subsection{Discussion}
\label{sec:perf_discussion}
The huge difference in training caused by using the original and the augmented training samples are very interesting since the most samples in the augmented training data are artificially generated using the original samples. Our hypothesis is that the context information of PCB components is well represented by the original samples due to the use of one-class training data, while the unpredictable variations of background is strongly suppressed. By augmenting the training samples, spatial transformations of objects, \eg. translation and rotation, are better represented in data. This leads to a more reliable feature extraction in CNNs. The successful training of CNNs using limited training samples relies on the fact that the diversity of data is significantly reduced using our proposed detection pipeline. Besides the proposed CNN-WRDF, the dropout of forwarded features also gives a positive push to the regression performance. By randomly partly deactivating features fed to fcs for localizing the object presented in the current region proposal, the regression function is forced to use redundant features for activating each neuron in fcs instead of using only one single or few features. In case of partly absent features of objects, especially low-level features, stable predictions can still be obtained using this redundancy-based mechanism.

\section{Conclusion}
\label{sec:conclusion}
As demonstrated on the PCB data set, our proposed detection pipeline can be applied without great adoption difficulties to improve the performance of existing algorithms by exploiting the power of DL. Using the one-class-data training strategy, the barrier cause by the requirement on a huge amount of reference data in the common cases is lifted and the CNNs for the region regression purpose can be successfully trained using samples of moderate size. Thanks to the novel CNN-WRDF architecture, it is possible to achieve superior regression results by slightly modifying some basis CNN architectures and training them using standard hyperparameters. Similar results were also obtained on Facial Keypoints Detection data set~\cite{kaggle2016}. This confirms the good accessibility of the proposed detection, even for the users without much experience with DL.


%
%



\bibliographystyle{IEEEtran}
\bibliography{IEEEfull,RecycledDeepFeatures_WL_Arxiv}
%
%
%

\end{document}